\documentclass[10pt,twocolumn,letterpaper]{article}

\usepackage[pagenumbers]{cvpr}   % To force page numbers for arXiv version
\usepackage{graphicx}
\usepackage{amsmath}
\usepackage{amssymb}
\usepackage{booktabs}
\usepackage{multirow}
\graphicspath{{./}{img/}}
\usepackage{algorithm}
\usepackage{algorithmic}

\usepackage[accsupp]{axessibility} % Improves PDF readability for those with visual impairments.

\usepackage[pagebackref=true,breaklinks=true,colorlinks=true,bookmarks=false,linkcolor={red!50!black},urlcolor={darkgray!80!black},citecolor={green!50!black}]{hyperref} 

\usepackage[capitalize]{cleveref}
\crefname{section}{Sec.}{Secs.}
\Crefname{section}{Section}{Sections}
\Crefname{table}{Table}{Tables}
\crefname{table}{Tab.}{Tabs.}

\begin{document}

%%%%%%%%% TITLE
\title{SVLL: Staged Vision-Language Learning for \\ Physically Grounded Embodied Task Planning}

%%%%%%%%% AUTHORS
\author{
Yuyuan Yang$^{1}\footnotemark[1]$ \enspace
Junkun Hong$^{4}\footnotemark[1]$ \enspace
Hongrong Wang$^{3}\footnotemark[1]$ \enspace
Honghao Cai$^{1,3}$ \enspace
Xunpeng Ren$^{5}$ \vspace{0.1em} \\
Ge Wang$^{1,2}$ \enspace
Mingcong Lei$^{1}$ \enspace
Shenhao Yan$^{1}$ \enspace
Jiahao Yang$^{1}$ \enspace
Chengsi Yao$^{1}$ \vspace{0.1em} \\
Xi Li$^{1}$ \enspace
Yiming Zhao$^{1,6,7}$ \enspace
Yatong Han$^{1,7}$ \enspace
Jinke Ren$^{1,2}\footnotemark[2]$ \vspace{0.5em} \\
{\small $^1$FNii-Shenzhen, CUHKSZ \quad $^2$SSE, CUHKSZ \quad $^3$SAI, CUHKSZ} \vspace{0.1em} \\
{\small $^4$Shenzhen University \quad $^5$University of Sydney \quad $^6$Harbin Engineering University \quad $^7$Ising AI}
}

\maketitle

\begingroup
\renewcommand{\thefootnote}{\fnsymbol{footnote}}
\footnotetext[1]{Equal contribution.}
\footnotetext[2]{Corresponding author: \href{mailto:jinkeren@cuhk.edu.cn}{\textcolor{magenta}{jinkeren@cuhk.edu.cn}}.}
\endgroup
%%%%%%%%% ABSTRACT
\begin{abstract}
Embodied task planning demands vision-language models to generate action sequences that are both visually grounded and causally coherent over time. However, existing training paradigms face a critical trade-off: joint end-to-end training often leads to premature temporal binding, while standard reinforcement learning methods suffer from optimization instability. To bridge this gap, we present Staged Vision-Language Learning (SVLL), a unified three-stage framework for robust, physically-grounded embodied planning. In the first two stages, SVLL decouples spatial grounding from temporal reasoning, establishing robust visual dependency before introducing sequential action history. In the final stage, we identify a key limitation of standard Direct Preference Optimization (DPO), its purely relative nature---optimizing only the preference gap between winning and losing trajectories while neglecting absolute likelihood constraints on optimal path, often yields unsafe or hallucinated behaviors. To address this, we further introduce Bias-DPO, a novel alignment objective that injects an inductive bias toward expert trajectories by explicitly maximizing likelihood on ground-truth actions while penalizing overconfident hallucinations. By anchoring the policy to the expert manifold and mitigating causal misalignment, SVLL, powered by Bias-DPO, ensures strict adherence to environmental affordances and effectively suppresses physically impossible shortcuts. Finally, extensive experiments on the interactive AI2-THOR benchmark and real-world robotic deployments demonstrate that SVLL outperforms both state-of-the-art open-source (e.g., Qwen2.5-VL-7B) and closed-source models (e.g., GPT-4o, Gemini-2.0-flash) in task success rate, while significantly reducing physical constraint violations.
\end{abstract}

%%%%%%%%% BODY TEXT
\section{Introduction}
\label{sec:intro}
\begin{figure}[t]
  \centering
  \includegraphics[width=\linewidth]{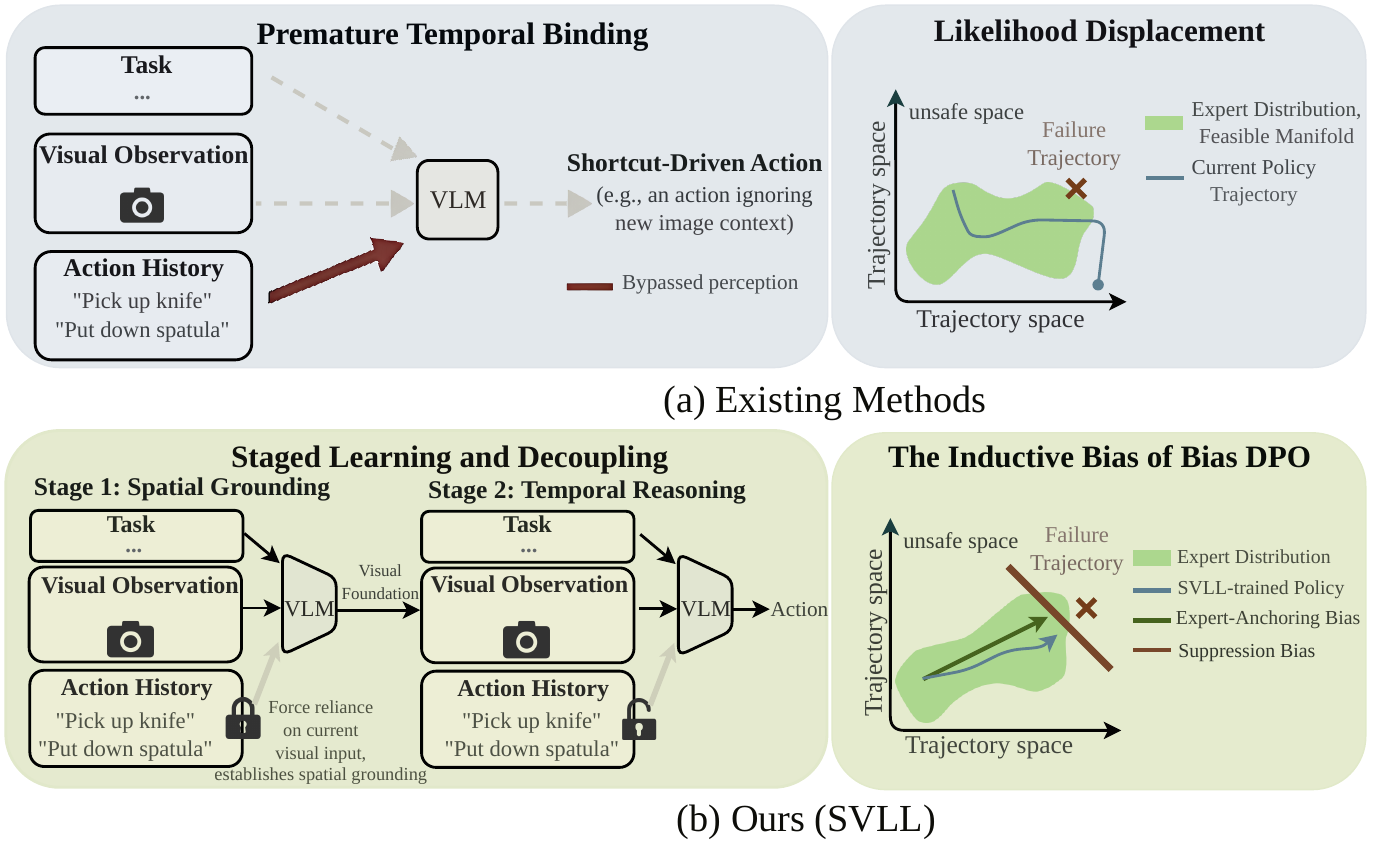}
  \caption{\textbf{A comparison between existing task planning methods and our proposed SVLL framework.} \textbf{(a) Existing Methods:} Existing methods suffer from premature temporal binding due to joint end-to-end training and likelihood displacement caused by standard DPO, which pushes the policy away from the expert distribution. \textbf{(b) Ours (SVLL):} SVLL addresses these issues by integrating staged learning and Bias-DPO, where the former decouples spatial grounding from temporal reasoning, while the latter introduces strong inductive biases to explicitly anchor the policy to the feasible expert manifold and suppresses confident errors.}
  \label{fig:teaser}
\end{figure}

Embodied task planning is a fundamental challenge in robotics, requiring the ability to translate high-level natural language instructions into executable, temporally coherent subtasks~\cite{shi2025hirobotopenendedinstruction,zhang2025hirtenhancingroboticcontrol,wu2023embodiedtaskplanninglarge}. Recent advances in large vision-language models (VLMs) have demonstrated promising capabilities in this domain, with some models even supporting extremely long context windows~\cite{bai2025qwen25vltechnicalreport,comanici2025gemini25pushingfrontier,deepseekai2025deepseekv3technicalreport}. Consequently, several works have leveraged pre-designed prompting strategies to guide VLMs toward task completion~\cite{fu2024what,shin2025socraticplannerselfqabasedzeroshot,rana2023sayplangroundinglargelanguage,hu2023lookleapunveilingpower,kim2024contextawareplanningenvironmentawarememory,singh2022progpromptgeneratingsituatedrobot}. However, direct deployment of state-of-the-art VLMs, such as Qwen2.5-72B-Instruction~\cite{wallace2025efficacyimagesimilaritymetric}, in dynamic embodied interactive tasks suffers from significant limitations. 
%When faced with dynamic, changing environments, 
These models often generate physically inconsistent actions due to a profound lack of grounded, real-time scene understanding~\cite{ma2026surveyvisionlanguageactionmodelsembodied}. 
For instance, when instructed to ``heat the apple in the microwave'', a standard VLM may correctly navigate to the appliance; however, if the microwave is already open, the model may ignore the visual cue and rigidly output an \texttt{Open(Microwave)} action, resulting in an invalid physical interaction. Furthermore, once the environment state changes mid-task, these models frequently fall into repetitive execution loops---such as outputting \texttt{Navigate(Microwave)} even after successfully arriving at the target. This occurs because they overfit to textual action histories rather than grounding next decisions in the updated visual observations.

As illustrated in \Cref{fig:teaser}, the aforementioned failures stem from two fundamental flaws in current training paradigms: 1) {\textit{premature temporal binding in joint training}}: Existing methods predominantly rely on joint Supervised Fine-Tuning (SFT) to train models on full trajectory contexts~\cite{wu2023embodiedtaskplanninglarge,ji2025robobrainunifiedbrainmodel,chen2024robogptintelligentagentmaking,brohan2023rt2visionlanguageactionmodelstransfer}. By exposing models simultaneously to visual observations and historical action sequences, these approaches induce premature temporal binding, i.e., the model learns to shortcut reasoning by relying on textual patterns rather than causally grounding decisions in the current visual state. This leads to brittle behavior under environmental dynamics, e.g., repeating past actions despite state changes~\cite{chen2025densejumpflowmatchingnonuniform}; 2) {\textit{causal misalignment in reasoning and policy alignment}: To force functional correctness, existing methods typically employ either pre-written system-level Chain-of-Thought (CoT) prompts~\cite{wang2025auxthinkexploringreasoningstrategies,fu2024what,shin2025socraticplannerselfqabasedzeroshot,rana2023sayplangroundinglargelanguage,hu2023lookleapunveilingpower,kim2024contextawareplanningenvironmentawarememory,singh2022progpromptgeneratingsituatedrobot,kojima2023largelanguagemodelszeroshot} or complex reward functions and standard preference optimization (e.g., Direct Preference Optimization (DPO))~\cite{song2024trialerrorexplorationbasedtrajectory,wang2025worldmodelingmakesbetter,ma2024eurekahumanlevelrewarddesign,wang2024rlvlmfreinforcementlearningvision,yao2026rewardevolutiongraphofthoughtsbilevel,schulman2017proximalpolicyoptimizationalgorithms,ma2023eureka,shao2024deepseekmathpushinglimitsmathematical}. However, these approaches suffer from a severe causal misalignment. 
%In standard text generation, lengthy textual reasoning or optimizing relative preference---simply pushing the probability of a ``winning'' sequence higher than a ``losing'' one---is often sufficient. However, embodied planning is governed by strict, absolute physical prerequisites (e.g., a microwave door must be physically open before an object can be placed inside). When these standard techniques are applied to the embodied domain, the model often learns statistical textual shortcuts rather than true physical causality. Instead of recognizing the current visual state as the absolute precondition for the next action, the policy merely memorizes that certain reasoning strings or action tokens are statistically safer to output given the context. This causal misalignment prevents the model from anchoring to the narrow, physically feasible expert manifold, inevitably resulting in ``high-confidence illusions''---actions that are statistically favored by the text but physically impossible in the active environment.
While relative preference optimization suffices for standard text generation, embodied planning is governed by strict, absolute physical constraints. Directly applying standard alignment methods in embodied domains causes models to exploit statistical textual shortcuts, rather than grounding decisions in visual-physical causality. As a result, the policy fails to anchor to the physically feasible expert manifold, inevitably producing high-confidence illusions---actions that are textually favored, yet physically invalid in the current environment.

\begin{figure*}[t]
  \centering
  \includegraphics[width=0.9\textwidth]{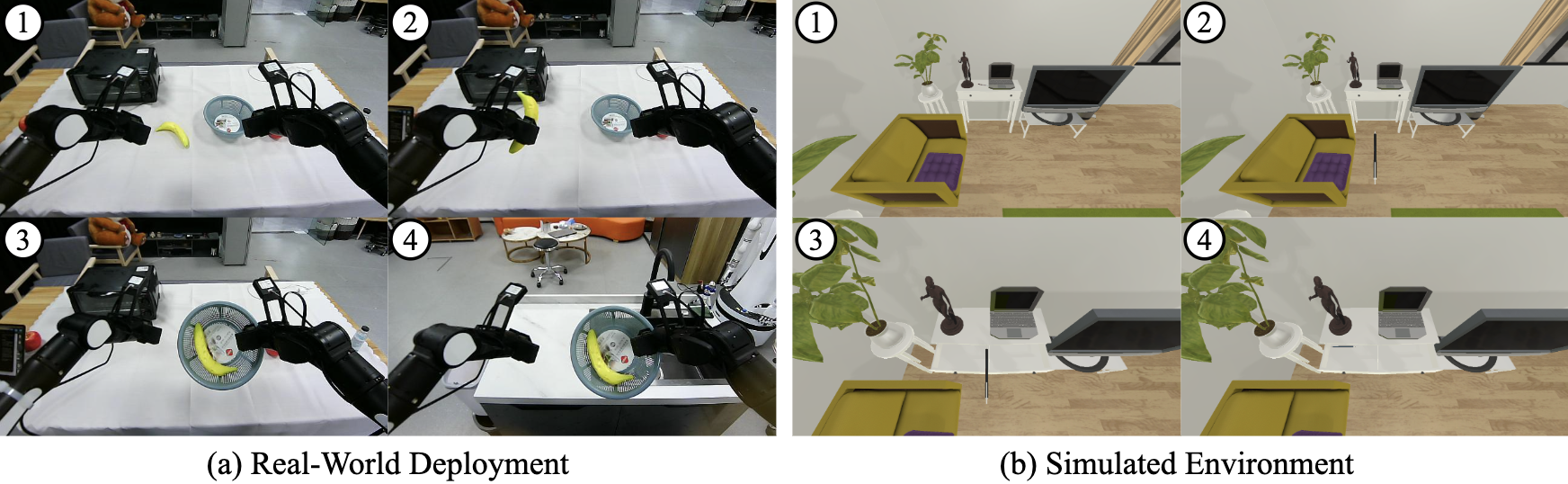}
  \caption{\textbf{Execution of long-horizon embodied tasks via SVLL.} \textbf{(a) Real-World Deployment:} A physical robotic arm executes a sequence of visually grounded actions: locating a banana, picking it up, placing it into a basket, and moving the basket to a designated target. \textbf{(b) Simulated Environment:} An agent in an interactive 3D simulator successfully performs a multi-step task involving finding a pen, picking it up, opening a drawer, and physically placing the pen inside. Both deployments demonstrate strict adherence to physical constraints and robust causal reasoning.}
  \label{fig:abstract_teaser}
\end{figure*}

To overcome these challenges, we propose Staged Vision-Language Learning (SVLL), a unified three-stage training framework specifically designed for the planner module in hierarchical vision-language architectures, aiming to address both temporal overfitting and likelihood displacement. Figure \ref{fig:abstract_teaser} illustrates the process of SVLL in executing long-horizon embodied tasks, which seamlessly grounds multi-step temporal reasoning into actionable, physics-compliant behaviors in physical environments. Our main contributions are summarized below:

\begin{enumerate}
  \item \textbf{Staged Decoupling Strategy:} We introduce a strict decoupling between spatial foundation mapping and temporal series reasoning in the first two stages of SVLL. Specifically, we force the planner to first identify functional affordances from static visual observations before introducing action history, thereby preventing the model from bypassing perception through historical shortcuts.

  \item \textbf{Bias-DPO-based Alignment Mechanism:} In the crucial third stage, we propose Bias-DPO (B-DPO), a novel alignment objective that transcends standard relative ranking. B-DPO introduces a strong inductive bias to explicitly amplify the probability mass of expert trajectories. By combining this inductive bias with a threshold-triggered penalty, B-DPO eliminates confident reasoning hallucinations while ensuring the policy remains strictly grounded in physical environments.

  \item \textbf{State-of-the-Art Performance and Safety:} Extensive experiments demonstrate that our SVLL framework (with 7B parameters) achieves a 78.35\% success rate on the highly interactive AI2-THOR benchmark, significantly outperforming closed-source models (e.g., Gemini-2.0-flash, GPT-4o) and domain-specific baselines. Crucially, in zero-shot real-world robotic deployments, the B-DPO method reduces constraint violations to 4.35\%, ensuring highly reliable and safe execution in physical environments.
\end{enumerate}

\section{Related Work}
\label{sec:related-work}
\subsection{Embodied Task Planning}
Embodied task planning aims to decompose high-level natural language instructions into a sequence of executable sub-tasks, enabling agents to complete tasks in complex environments~\cite{liang2023codepolicieslanguagemodel}. Early studies primarily employed frozen models to generate plans through various cue strategies and prompt engineering~\cite{huang2022innermonologueembodiedreasoning,ahn2022icanisay}. While current foundation models possess extensive context processing capabilities, their zero-shot performance remains unsatisfactory in complex, dynamically changing environments~\cite{yao2023reactsynergizingreasoningacting}. To address this issue, subsequent research has adopted SFT~\cite{wu2023embodiedtaskplanninglarge, ji2025robobrainunifiedbrainmodel,chen2024robogptintelligentagentmaking} to allow models to learn directly from expert demonstrations. 

Beyond SFT, recent studies have also employed preference learning and reinforcement learning methods to further align model behaviors. Methods based on Proximal Policy
Optimization (PPO)~\cite{schulman2017proximalpolicyoptimizationalgorithms} and Group Relative Policy Optimization (GRPO)~\cite{shao2024deepseekmathpushinglimitsmathematical} often rely on complex reward functions to train reward models, while DPO-based approaches~\cite{rafailov2024directpreferenceoptimizationlanguage} primarily focus on comparing successful expert trajectories with the agent's failure trajectories. However, applying standard preference learning to this domain often leads to a severe causal misalignment. The distributional bias between the extreme sparsity of successful samples and the infinite diversity of failure modes is frequently overlooked, fundamentally neglecting the inherent structural imbalance of the action space. To overcome this limitation, we propose a staged framework, in which the first stage establishes robust spatial grounding through observation mapping without historical information, while the second stage gradually introduces temporal causal dependencies. Finally, the causal misalignment in standard preference optimization is resolved using B-DPO, which structurally anchors the model to the expert distribution, thereby achieving robust planning without requiring manually designed reward signals.

\subsection{Vision-Language Model Reasoning}
The development of VLMs has shifted from passive perception tasks (e.g., visual question answering and image description) to active reasoning and decision-making~\cite{thawakar2025llamavo1rethinkingstepbystepvisual,wang2025enhancingreasoningabilitymultimodal,xu2025llavacotletvisionlanguage,xu2025redstardoesscalinglongcot,zhang2024improvevisionlanguagemodel,kimiteam2025kimik15scalingreinforcement,wang2024qwen2vlenhancingvisionlanguagemodels}. Early foundation models, such as PaLM-E~\cite{driess2023palmeembodiedmultimodallanguage}, have demonstrated that VLMs can implicitly encode robot states. Recent studies, such as CoT-VLA~\cite{zhao2025cotvlavisualchainofthoughtreasoning} and ReFineVLA~\cite{vanvo2025refinevlareasoningawareteacherguidedtransfer}, have directly integrated CoT reasoning into the vision domain~\cite{zhang2024multimodalchainofthoughtreasoninglanguage,mu2023embodiedgptvisionlanguagepretrainingembodied,zawalski2025roboticcontrolembodiedchainofthought}. However, the lack of a mechanism for continuously updating planning strategies through interactive feedback with the environment makes it difficult to adapt to dynamic perturbations in long-horizon embodied tasks~\cite{ahn2022icanisay,brohan2023rt2visionlanguageactionmodelstransfer,kim2024openvlaopensourcevisionlanguageactionmodel,Lin_2023}. To address this issue, our dataset leverages image-text interleaving to dynamically generate plans that adapt to dynamic interactive environments. This ensures that abstract reasoning capabilities align with the strict causal relationships inherent in embodied planning, rather than merely executing simple instructions, thus enabling robust functional planning.

\section{Method}
\label{sec:method}
\begin{figure}[t]
  \centering
  \includegraphics[width=\linewidth]{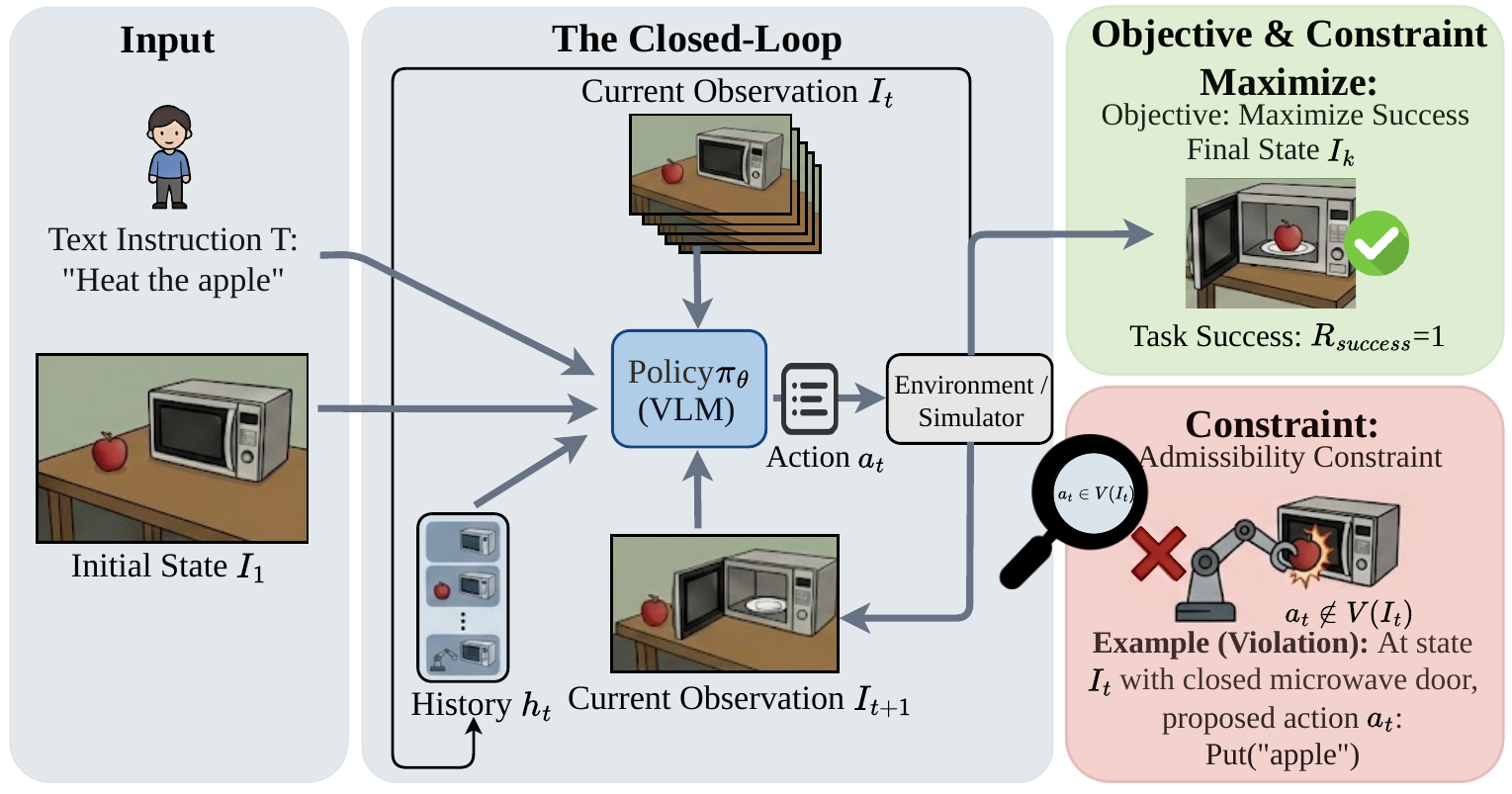}
  \caption{\textbf{Formulation of embodied task planning.} The agent engages in a closed-loop interaction with the environment based on natural language instructions. The objective is to maximize the task success rate ($\mathcal{R}_{\text{success}}$), subject to a critical admissibility constraint---every intermediate action $a_t$ must be physically executable and grounded within the valid visual affordance space $\mathcal{V}(I_t)$ of the current observation.}
  \label{fig:problem_formulation}
  \vspace{-8pt}
\end{figure}

\subsection{Problem Formulation}
As illustrated in \Cref{fig:problem_formulation}, embodied task planning can be formalized as a dynamic, closed-loop sequential decision-making process. Given a high-level natural language instruction $T$, the interaction proceeds in discrete time steps. At each step $t$, the agent perceives an egocentric visual observation $I_t$, which reflects the current state of the environment. The policy $\pi_\theta(a_t \mid I_t, T, h_t)$ is responsible for generating an executable function call $a_t$, where $h_t = [a_1, \dots, a_{t-1}]$ denotes the history of previously executed actions. Upon execution, the environment transitions to a new state, yielding an updated observation $I_{t+1}$. This iterative cycle continues until the agent outputs a termination token at step $K$. 

Formally, let $\tau = \{(I_1, a_1), (I_2, a_2), \dots, (I_K, a_K)\}$ denote a sampled trajectory. The objective is to identify optimal parameters $\theta$ that maximize the expected task success rate, under the strict constraint that every intermediate action is physically grounded in the active scene. This is formulated as the following optimization problem:
\begin{subequations} 
\begin{flalign}
    \max_{\theta} &~~J(\theta) = \mathbb{E}_{\tau \sim \pi_\theta} \left[ \mathcal{R}_{\text{success}}(I_K, T) \right] \\\text{s.t.}  &~~a_t \in \mathcal{V}(I_t), \; \forall t \in \{1, \dots, K\},
\end{flalign}
\end{subequations}
where $\mathcal{R}_{\text{success}}(I_K, T) \in \{0, 1\}$ is a binary reward indicating whether the final environment state $I_K$ satisfies the instruction $T$, and $\mathcal{V}(I_t)$ denotes the manifold of physically valid and executable actions given the visual affordances in $I_t$.

\subsection{SVLL Framework}
\label{sec:svll}
\begin{figure*}[t]
  \centering
  \includegraphics[width=0.95\textwidth]{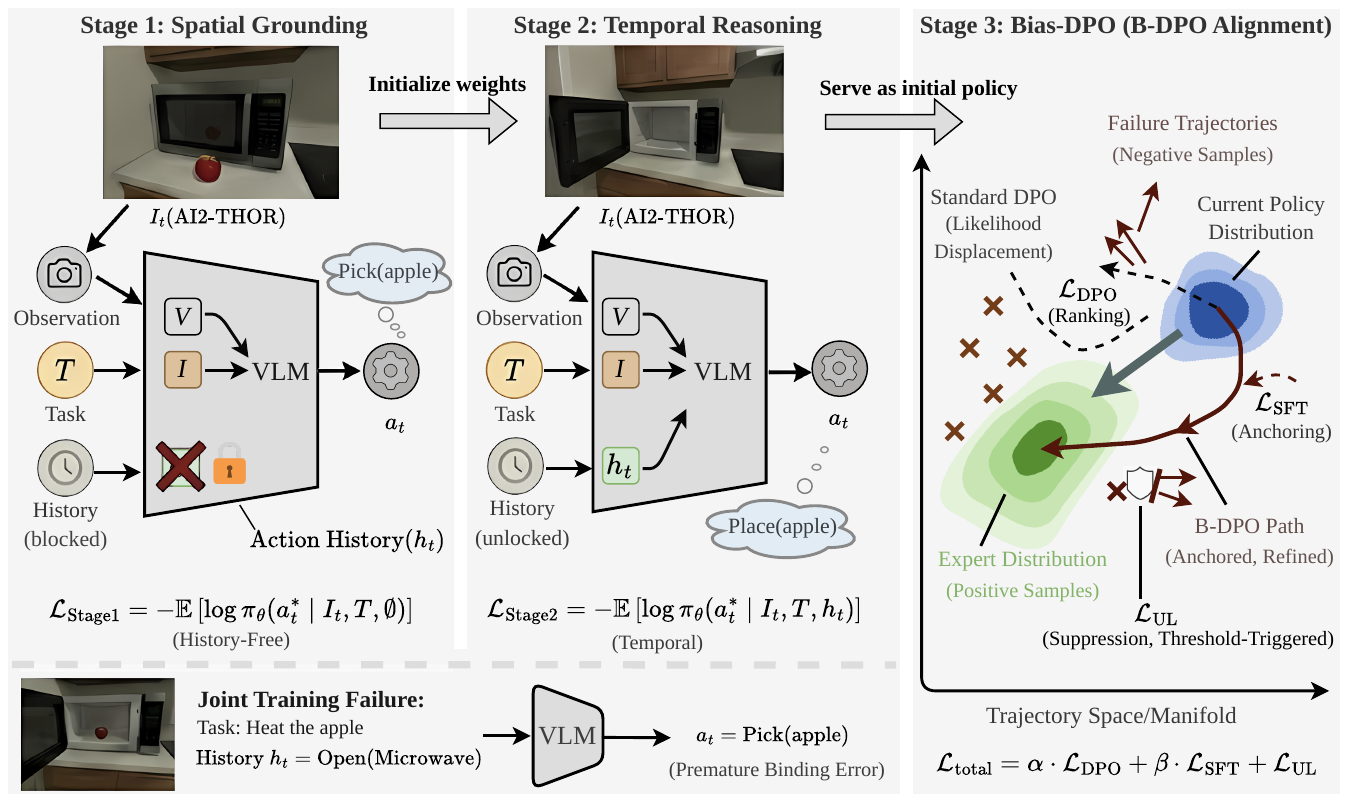}
  \caption{\textbf{Overview of the SVLL framework.} To mitigate the premature temporal binding in standard end-to-end training, SVLL decouples spatial and temporal learning across three stages: Stage 1 blocks the action history to force reliance on current visual affordances; Stage 2 unlocks the history context, inheriting robust visual features from Stage 1 to learn sequential dependencies; Stage 3 refines the initial policy by anchoring it to the expert manifold via an auxiliary $\mathcal{L}_{\text{SFT}}$ and a threshold-triggered penalty $\mathcal{L}_{\text{UL}}$, effectively overcoming the likelihood displacement inherent in standard DPO.}
  \label{fig:svll_overview}
\end{figure*}
Standard training paradigms typically optimize $\pi_\theta$ on the full trajectory context $(I, T, h)$ from scratch. However, we identify two critical failure modes inherent in this approach: 1) premature temporal binding, where the model prematurely overfits to textual history shortcuts during early training, compromising its grounding in visual perception; and 2) causal misalignment, where conventional relative alignment methods overlook the structural imbalance in the action space, thereby undermining the absolute stability of the expert policy. To systematically address these issues, we propose SVLL, a unified three-stage framework that explicitly decouples spatial and temporal learning before performing physics-constrained alignment. The overall framework is illustrated in \Cref{fig:svll_overview}.

\subsubsection{Stage 1: Spatial Grounding}
In the first stage, we enforce a strict focus on visual affordances to prevent premature temporal binding. To achieve this, we curate a specific subset from the embodied trajectory dataset introduced by Zhang et al.~\cite{zhang2025embodiedreasonersynergizingvisualsearch}. We rigorously filter the data samples to retain only samples in which the optimal next action can be determined solely from the current frame without requiring any historical context (e.g., reactive actions like \textit{pick} and \textit{place}). The model is trained to predict the optimal action $a^*_t$ while explicitly blocking the history channel, which is formulated as:
\begin{equation}
  \mathcal{L}_{\text{Stage1}} = -\mathbb{E}_{(I_t, T, a_t^*) \sim \mathcal{D}_{\text{spatial}}} \left[\log \pi_\theta(a_t^* \mid I_t, T, \emptyset)\right].
\end{equation}

During training, we freeze the vision encoder and perform full fine-tuning on the remaining multimodal architecture. By removing the temporal context $h_t$, the model is prevented from exploiting statistical correlations in textual history. Instead, it is compelled to resolve the instruction strictly by parsing the current pixel-space observation $I_t$. This stage lays a robust, perception-driven foundation for spatial understanding before any sequential bias is introduced.

\subsubsection{Stage 2: Temporal Reasoning}
Once spatial grounding is consolidated, we unlock the history context $h_t = [a_1, \dots, a_{t-1}]$ formatted from continuous trajectory sequences. Critically, the policy is initialized using the spatially-grounded checkpoint from Stage 1. With the vision encoder kept frozen, we proceed with full fine-tuning on sequential rollouts:
\begin{equation}
  \mathcal{L}_{\text{Stage2}} = -\mathbb{E}_{(I_t, T, h_t, a_t^*) \sim \mathcal{D}_{\text{temporal}}} \left[\log \pi_\theta(a_t^* \mid I_t, T, h_t)\right].
\end{equation}

In this stage, the model learns to integrate sequential dependencies, such as recalling previously explored areas or tracking multi-step sub-goals. Since spatial mapping capabilities were already established in Stage 1, the model learns to treat the continuous history $h_t$ as a supplementary temporal cue rather than a primary crutch for decision-making. By operating within this enriched context, the policy develops causal temporal reasoning while preserving the strict visual grounding capabilities acquired in Stage 1.

\subsubsection{Stage 3: Alignment via B-DPO}
In the final stage, we enforce strict adherence to physical constraints by freezing the base model and applying Low-Rank Adaptation (LoRA) to all linear projection layers. To overcome the likelihood displacement caused by standard relative optimization---where the absolute probability of the optimal path decays merely to maximize the preference margin---we introduce B-DPO, a novel preference optimization objective explicitly designed to rectify the structural imbalance of the action space. 

Specifically, given a prompt $x = (I, T, h)$, a chosen expert trajectory $y_w$, and a rejected trajectory $y_l$, B-DPO enhances the standard DPO formulation through three key steps: First, an internal positive weight ($w > 1$) is applied to amplify the margin for chosen actions, reinforcing the model's alignment with expert behavior. Second, a strong inductive bias is injected via an auxiliary supervised fine-tuning loss $\mathcal{L}_{\text{SFT}}$, which anchors the model to the feasible expert distribution and explicitly prevents the aforementioned likelihood displacement; Third, overconfident hallucinations are actively suppressed through a threshold-triggered unlikelihood penalty $\mathcal{L}_{\text{UL}}$, which activates only when the likelihood of the rejected trajectory exceeds a predefined safety threshold $\tau$, thereby preserving model conservatism. Based on the three steps, the total loss is defined as
\begin{equation}
  \mathcal{L}_{\text{total}} = a \cdot \mathcal{L}_{\text{DPO}} + b \cdot \mathcal{L}_{\text{SFT}} + \mathcal{L}_{\text{UL}},
\end{equation}
where $a$ and $b$ are two weight coefficients.

Overall, the detailed implementation, including label smoothing ($\varepsilon$) and threshold suppression ($\tau$), is provided  in Algorithm~\ref{alg:bdpo}.

\begin{algorithm}[t]
\caption{Bias Direct Preference Optimization}
\label{alg:bdpo}
\textbf{Hyperparameters:} KL temperature $\beta$, chosen weight $w$, DPO weight $a$, SFT weight $b$, unlikelihood threshold $\tau$, label smoothing $\varepsilon$.\\
\textbf{Input:} Trainable policy $\pi_\theta$, reference policy $\pi_{\text{ref}}$, prompt $x$, chosen response $y_w$, rejected response $y_l$. \\
\textbf{Output:} The total loss $\mathcal{L}_{\text{total}}$ used for backpropagation.
\begin{algorithmic}[1]
    \STATE $\text{chosen\_logratio} \gets \log \pi_\theta(y_w \mid x) - \log \pi_{\text{ref}}(y_w \mid x)$. \hfill // \textit{Calculate log-ratios}
    \STATE $\text{rejected\_logratio} \gets \log \pi_\theta(y_l \mid x) - \log \pi_{\text{ref}}(y_l \mid x)$.
    
    \STATE $\text{logits} \gets w \cdot \text{chosen\_logratio} - \text{rejected\_logratio}$. \hfill // \textit{Core B-DPO logits ($w > 1$)}
    
    \STATE $\mathcal{L}_{\text{DPO}} \gets -(1 - \varepsilon) \log \sigma(\beta \cdot \text{logits}) - \varepsilon \log \sigma(-\beta \cdot \text{logits})$. \hfill // \textit{Base DPO loss}
    
    \STATE $\mathcal{L}_{\text{SFT}} \gets -\log \pi_\theta(y_w \mid x)$. \hfill // \textit{SFT auxiliary loss}
    
    \STATE $\mathcal{L}_{\text{UL}} \gets \max(0, \log \pi_\theta(y_l \mid x) - \tau)$. \hfill // \textit{Unlikelihood penalty}
    
    \STATE $\mathcal{L}_{\text{total}} \gets a \cdot \mathcal{L}_{\text{DPO}} + b \cdot \mathcal{L}_{\text{SFT}} + \mathcal{L}_{\text{UL}}$. \hfill // \textit{Combine total loss}
    \RETURN $\mathcal{L}_{\text{total}}$
\end{algorithmic}
\end{algorithm}

\section{Experiments}
\label{sec:experiments}
In this section, we present a comprehensive experimental evaluation of proposed SVLL framework, aiming to answer the following three key questions:
\begin{itemize}
    \item \textbf{Q1:} How does our SVLL framework perform comparing to state-of-the-art closed-source and open-source foundation models in embodied task planning?
    \item \textbf{Q2:} Does the decoupled SFT (Stage 1 + Stage 2) effectively mitigate premature temporal binding compared to conventional end-to-end training?
    \item \textbf{Q3:} How effectively does B-DPO (Stage 3) suppress overconfident reasoning hallucinations and physical constraint violations?
\end{itemize}

\subsection{Experimental Setup}
\textbf{Dataset and Environment.} 
We conduct experiments using the embodied trajectory dataset introduced by Zhang et al.~\cite{zhang2025embodiedreasonersynergizingvisualsearch}, which is built upon the highly interactive AI2-THOR simulator. This dataset is rigorously constructed by collecting expert demonstration rollouts across diverse indoor environments. For evaluation, we construct our benchmark by sampling from its designated test set. This test set comprehensively covers a wide spectrum of interactive task types with varying complexities, including fundamental state changes, single-object manipulation, multi-object compositional tasks, visual search, and complex sequential reasoning. We systematically sample representative episodes from each of these categories to rigorously evaluate the models. To adapt these continuous trajectories for VLM training, raw demonstrations are processed into interleaved multimodal sequences. Specifically, each data point is formatted as a tuple $(I_t, T, h_t, a_t)$, where $I_t$ denotes the current egocentric visual observation, $T$ is the high-level natural language instruction (e.g., ``Would you mind helping me place the Pen in the Drawer?''), $h_t$ is the explicit action history up to the current step, and $a_t$ is the ground-truth executable next action (e.g., \texttt{Navigate(CoffeeTable)}).

\noindent\textbf{B-DPO Preference Data Construction.}
To train the alignment model in Stage 3, we dynamically construct a near-policy preference dataset.  The Stage 2 policy is deployed back into the environment using temperature sampling to promote diverse exploratory behaviors, from which we collect trajectory pairs $(y_w, y_l)$. To explicitly address the likelihood displacement problem, we sampled 1,000 distinct negative trajectories $y_l$ that represent typical embodied hallucinations. These negative samples are strictly categorized into four types of physical constraint violations:
{\textit{1) Invalid Navigation:} The agent has already reached the target but outputs \texttt{Navigate} rather than \texttt{Pick}.
{\textit{2) Repetitive Search:} The agent has visually located the target object but executes another redundant \texttt{Navigate} action.
{\textit{3) Erroneous Navigation:} The agent is already holding the target object at the target receptacle but outputs \texttt{Navigate} instead of \texttt{Place}.
{\textit{4) Failure to Terminate:} The task is functionally completed, but the agent fails to output the \texttt{TaskCompleted} token.

\noindent\textbf{Implementation Details.}
We employ the Qwen2.5-VL-7B~\cite{bai2025qwen25vltechnicalreport} model as our backbone architecture. The training process strictly follows the proposed three-stage SVLL framework using 8$\times$ NVIDIA A100 (80GB) GPUs. Stage 1 (History-Free Full Fine-Tuning) and Stage 2 (Full-Context Full Fine-Tuning) each require approximately 25 hours of training. In Stage 3, we freeze the base model parameters and apply LoRA to optimize the B-DPO objective within 4 hours.

\noindent\textbf{Baselines.}
We conduct a comprehensive evaluation of our framework against strong baselines across three distinct categories:
\begin{itemize}
    \item \textbf{Closed-Source VLMs:} Gemini-2.0-flash and GPT-4o~\cite{openai2024gpt4technicalreport}, serving as the upper bounds of zero-shot general reasoning capabilities.
    \item \textbf{Open-Source General VLMs:} The base Qwen2.5-VL-7B model~\cite{bai2025qwen25vltechnicalreport}, evaluated in zero-shot mode.
    \item \textbf{Embodied-Specific Models:} RoboBrain2.0-32B~\cite{ji2025robobrainunifiedbrainmodel} and the intermediate models of SVLL, namely SVLL-Stage 1 and SVLL-Stage 2.
\end{itemize}

\noindent\textbf{Evaluation Metrics.}
We evaluate the performance of our framework using a comprehensive set of metrics designed to capture both overall task success and fine-grained behavioral modes:
\begin{itemize}
    \item \textbf{Success Rate (SR):} The primary macroscopic metric, defined as the percentage of episodes in which the agent successfully completes all instructions and explicitly outputs the \texttt{TaskCompleted} token at the final state.
    \item \textbf{Success weighted by Path Length (SPL):} A standard metric for embodied efficiency, computed as $\text{SPL} = \frac{1}{N} \sum_{i=1}^{N} S_i \frac{l_i}{\max(p_i, l_i)}$, where $S_i$ is the binary success indicator for episode $i$, $p_i$ denotes the actual number of steps taken by the agent, and $l_i$ is the length of the shortest path from the expert demonstration.
    \item \textbf{Valid Action Rate (VAR):} A metric assessing spatial grounding capability. It measures the proportion of generated actions that are physically executable in the current environment state (e.g., predicting \texttt{Open(Microwave)} when no microwave is visible is classified as invalid). This metric directly reflects the visual understanding proficiency acquired in Stage 1.
    \item \textbf{Loop Rate (LR):} A diagnostic indicator of temporal overfitting, defined as the percentage of episodes in which the agent falls into repetitive action loops (e.g., oscillating between two states for $K$ consecutive steps or stalling by repeatedly issuing invalid actions). This metric validates the necessity of our decoupled training strategy in Stages 1 and 2.
    \item \textbf{Constraint Violation Rate (CVR):} A metric quantifying confident physical hallucinations, which measures the proportion of actions that trigger the four predefined physical logic errors defined in our B-DPO negative samples (e.g., continuing to search while already holding the target object, or failing to terminate upon completion). CVR explicitly evaluate the alignment efficacy of Stage 3.
\end{itemize}

\subsection{Main Results}
\begin{table*}[t]
\centering
\caption{\textbf{Main results on the AI2-THOR embodied planning benchmark.} We compare our SVLL framework (with 7B parameters) against closed-source VLMs, open-source general VLMs, and embodied-specific models. SR and SPL measure overall task performance (higher is better). VAR, LR, and CVR diagnose specific behavioral modes (VAR: higher is better; LR and CVR: lower is better). All metrics are reported in percentages (\%). Our full framework (SVLL-Stage 3) achieves state-of-the-art performance while drastically reducing logical hallucinations.}
\label{tab:main_results}
\begin{tabular}{ll ccccc}
\toprule
\textbf{Category} & \textbf{Model} & \textbf{SR} $\uparrow$ & \textbf{SPL} $\uparrow$ & \textbf{VAR} $\uparrow$ & \textbf{LR} $\downarrow$ & \textbf{CVR} $\downarrow$ \\
\midrule
\multirow{2}{*}{Closed-Source VLMs} 
& GPT-4o~\cite{openai2024gpt4technicalreport} & 48.45 & 46.83 & 59.68 & \textbf{1.03} & 40.32 \\
& Gemini-2.0-flash & 58.76 & 49.32 & 68.45 & 3.09 & 31.55 \\
\midrule
\multirow{1}{*}{Open-Source General VLMs} 
& Qwen2.5-VL-7B~\cite{bai2025qwen25vltechnicalreport} & 18.56 & 17.85 & 45.13 & 5.15 & 54.87 \\
\midrule
\multirow{5}{*}{\begin{tabular}[c]{@{}l@{}}Embodied-Specific \\ (Ours \& Baselines)\end{tabular}} 
& RoboBrain2.0-32B~\cite{ji2025robobrainunifiedbrainmodel} & 27.84 & 26.98 & 47.39 & 6.19 & 52.61 \\
\cmidrule{2-7}
& Ours (Direct Stage 2) & 68.04 & 55.24 & 78.37 & 10.31 & 30.65 \\
& Ours (SVLL-Stage 1) & 64.95 & 51.18 & 76.42 & 12.37 & 33.42 \\
& Ours (SVLL-Stage 2) & 74.23 & 62.29 & 84.16 & 8.25 & 29.18 \\
& \textbf{Ours (SVLL-Stage 3)} & \textbf{78.35} & \textbf{68.53} & \textbf{89.64} & 7.22 & \textbf{26.34} \\
\bottomrule
\end{tabular}
\vspace{-10pt}
\end{table*}
\label{subsec:main_results}
To address Q1, we conduct comprehensive evaluations comparing our framework against leading closed-source and open-source foundation models. The results are presented in Table~\ref{tab:main_results}. It is observed that our full SVLL-Stage 3 framework achieves state-of-the-art performance across most metrics, conclusively answering RQ1. While closed-source foundation models such as Gemini-2.0-flash and GPT-4o demonstrate strong general reasoning capabilities—achieving SR of 58.76\% and 48.45\% respectively—their relatively high CVR of 31.55\% and 40.32\% reveal a limitation in strict physical grounding. In permissive simulated environments, these high-capacity models often reach the final goal through iterative environmental feedback, inadvertently accumulating physically invalid actions, such as attempting to interact with occluded objects. 

Similarly, the open-source foundation model Qwen2.5-VL and the embodied-specific RoboBrain2.0-32B exhibit CVRs exceeding 52\%, highlighting the inherent challenge of translating visual reasoning into physically executable, grounded action sequences. In contrast, by explicitly aligning the model with physical constraints via the B-DPO objective, our 7B-parameter SVLL-Stage 3 model not only achieves the highest SR (78.35\%) but also attains the lowest CVR (26.34\%) and the highest VAR (89.64\%). Furthermore, its superior SPL (68.53\%) confirms that the agent completes tasks efficiently, following expert-like, causally coherent step sequences. These results demonstrate that a targeted, multi-stage structural alignment strategy can effectively bridge the physical grounding gap without solely relying on massive parameter scaling.

\subsection{Real-World Robot Deployment}
\label{subsec:real_world}

To demonstrate the practical viability and zero-shot transferability of our framework, we deploy the SVLL-Stage 3 policy onto a real-world robotic platform. 

\noindent\textbf{Hardware Setup and Tasks.} 
The real-world experiments are conducted using a 6-Degrees of Freedom (DoF) robotic manipulator equipped with a parallel-jaw gripper and a wrist-mounted RGB camera for egocentric vision perception. We evaluate the models on a rigorous set of 9 distinct real-world tasks, specifically designed to test compositional generalization. Each task requires the agent to execute long-horizon plans by sequentially combining fundamental primitive skills: \texttt{Navigate}, \texttt{Pick}, \texttt{Place}, \texttt{Open}, and \texttt{Close}. For instance, a single task sequence may require the agent to navigate to a target receptacle, explicitly open it, pick up an object, and physically place it inside, as visualized in \Cref{fig:abstract_teaser}.

\noindent\textbf{Baselines and Metrics.}
Given the high execution cost and stringent safety requirements of physical deployment, we benchmark our 7B-parameter model against the open-weight embodied foundation model: RoboBrain2.0-32B~\cite{ji2025robobrainunifiedbrainmodel}. Performance is evaluated using SR across the 9 tasks and CVR. In the physical domain, CVR serves as a critical safety metric, measuring the frequency of causally inconsistent or physically unexecutable commands (e.g., attempting to \texttt{Place} an object into a closed drawer without first issuing \texttt{Open}, or attempting to \texttt{Pick} an object that is occluded from view).

\begin{table}[t]
\centering
\caption{\textbf{Real-world zero-shot evaluation.} Comparison between the 32B-parameter state-of-the-art embodied model and our SVLL framework (with 7B parameters) across 9 real-world tasks. All metrics are reported in percentages (\%). Our SVLL-Stage 3 framework achieves superior SR while maintaining a significantly lower CVR, demonstrating robust causal reasoning and physical grounding capabilities.}
\label{tab:real_world_results}
\begin{tabular}{lccc}
\toprule
\textbf{Model} & \textbf{Parameters} & \textbf{SR} $\uparrow$ & \textbf{CVR} $\downarrow$ \\
\midrule
RoboBrain2.0~\cite{ji2025robobrainunifiedbrainmodel} & 32B & 22.22 & 10.94 \\
\midrule
\textbf{Ours (SVLL-Stage 3)} & \textbf{7B} & \textbf{55.56} & \textbf{4.35} \\
\bottomrule
\end{tabular}
\vspace{-8pt}
\end{table}

\noindent\textbf{Results.}
As shown in Table~\ref{tab:real_world_results}, our SVLL framework demonstrates exceptional zero-shot transfer capabilities. Despite having approximately one-fifth the parameter count of RoboBrain2.0-32B, our framework successfully completes 5 out of 9 tasks, achieving a highly competitive SR. As illustrated in \Cref{fig:real_world_comparison}, while the 32B baseline occasionally generates physically infeasible action sequences (e.g., skipping prerequisite state-changing actions like \texttt{Open} before attempting to  \texttt{Place} an object), our B-DPO optimized policy strictly grounds its decisions in valid physical affordances, thereby preventing catastrophic failures in the real-world execution.

\begin{figure*}[t]
  \centering
  \includegraphics[width=0.9\textwidth]{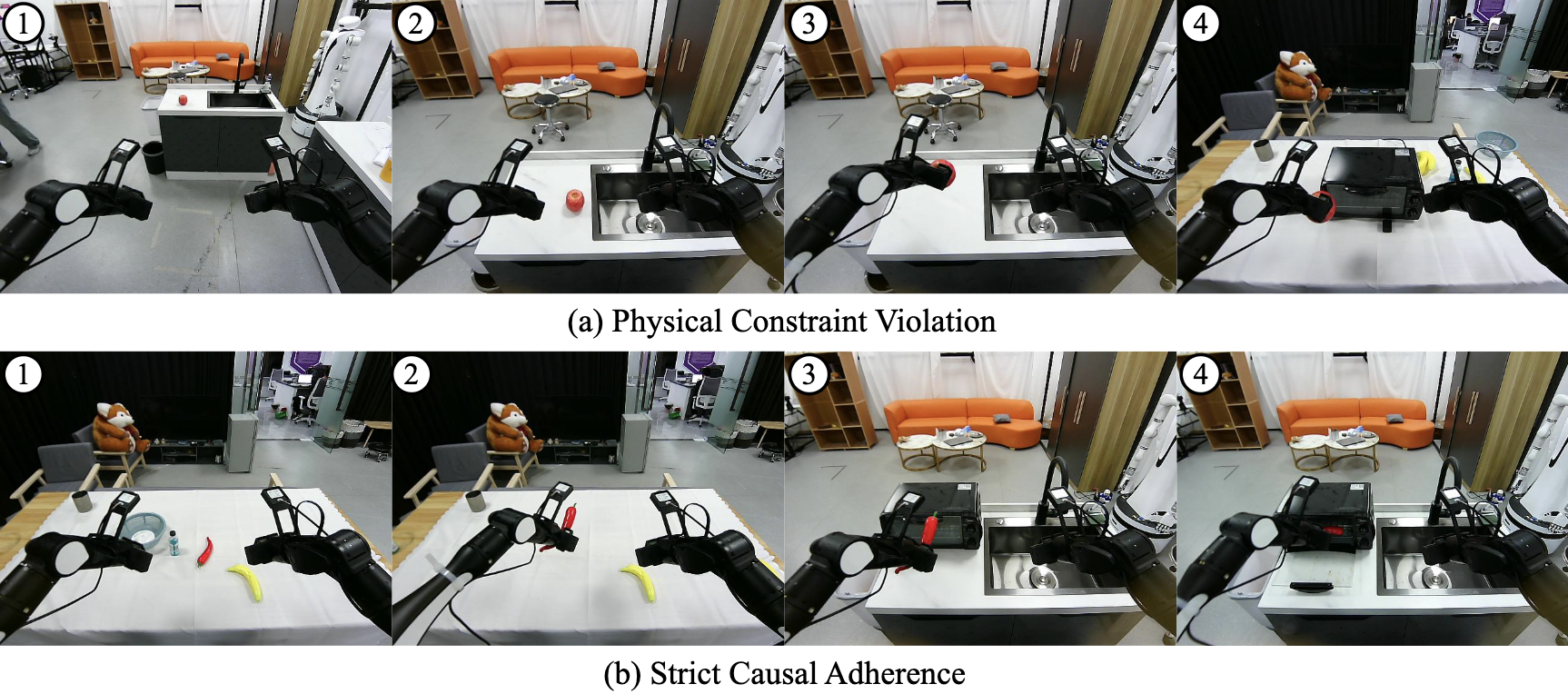}
  \caption{\textbf{Comparison of the RoboBrain2.0-32B model and our 7B-parameter SVLL-Stage 3 model in executing real-world tasks}. Both models are tasked with locating an object and placing it inside the microwave. \textbf{(a) Physical Constraint Violation:} The baseline model successfully navigates to and picks up the apple but commits a critical physical constraint violation by attempting to \texttt{Place} it without first opening the microwave door. \textbf{(b) Strict Causal Adherence:} Our model successfully locates the pepper, explicitly executes the prerequisite \texttt{Open} action, and then safely places the object inside, demonstrating strict adherence to causal physical constraints throughout the task.}
  \label{fig:real_world_comparison}
\end{figure*}

\subsection{Ablation Study on Staged Decoupling}
To demonstrate the necessity of decoupling spatial grounding from temporal reasoning, we compare our SVLL-Stage 2 model against a ``Direct Stage 2'' baseline, which skips Stage 1 entirely and applies standard joint SFT using the full action history $h_t$ from the outset. The results are shown in Table~\ref{tab:main_results}. We can observe that the Direct Stage 2 model exhibits a strong tendency to over-rely on temporal patterns. By gaining immediate access to $h_t$ during early fine-tuning, it learns to utilize textual sequence shortcuts, thereby reducing its attention to the current visual input $I_t$. Consequently, when deployed in dynamic environments where physical states deviate from expected trajectories, the model exhibits a high LR of 10.31\%, as it repetitively executes historical action sequences. 

In contrast, our decoupled framework effectively mitigates this modality imbalance. In SVLL-Stage 1, the model is trained exclusively on visual observations without any historical context, compelling it to develop robust spatial grounding. While this lack of temporal memory naturally results in a higher LR (12.37\%) due to limited contextual tracking, it establishes a critical foundation for visual dependency. When temporal history is subsequently reintroduced in SVLL-Stage 2, the model integrates sequential memory without becoming over-reliance on it. This structural prior yields significant performance gains: the VAR increases to 84.16\%, the SR improves to 74.23\%, and the LR is reduced to 8.25\%. These empirical results validate that establishing visual spatial grounding prior to temporal reasoning is a crucial design principle for achieving balanced modality utilization in multimodal embodied agents.

\subsection{Ablation Study on B-DPO Alignment}
While SVLL-Stage 2 achieves a strong SR of 74.23\%, standard SFT lacks explicit negative constraints, leaving the agent vulnerable to out-of-distribution physical states. This results in a CVR of 29.18\% due to physically infeasible commands. 

To address this, the Stage 3 in SVLL introduces B-DPO, which actively penalizes predefined categorized physical hallucinations. As shown in Table~\ref{tab:main_results}, this targeted alignment mechanism serves as a critical structural safeguard, reducing the CVR to 26.34\% while increasing the VAR to 89.64\%. Crucially, unlike standard DPO that induces likelihood displacement away from the expert distribution, our B-DPO objective structurally anchors the policy to the manifold of successful, physically-grounded trajectories. This mechanism effectively suppresses ``confident errors'' while preserving generative fidelity, ultimately boosting the overall SR to a state-of-the-art 78.35\% and reducing the LR to just 7.22\%. These results confirm that explicit, structured preference alignment is essential for causally coherent and physically reliable embodied task planning.

\section{Discussion and Limitations}
\label{sec:discussion}
Our findings challenge the prevailing assumption that preference optimization is fundamentally a ranking problem. While treating policy learning as relative preference ranking (e.g., standard DPO) proves effective in open-ended text generation, it fails in embodied intelligence, where valid behaviors are strictly governed by physical laws and causal structure. We observe that purely relative optimization often induces likelihood displacement: the model learns to dislike negative samples without learning to reproduce or adhere to the positive one. B-DPO addresses this by injecting a structural inductive bias, explicitly anchoring the policy to the expert distribution while surgically suppressing confident hallucinations. This ensures the agent confined to the sparse, physically feasible manifold of the task.

Despite these improvements, our framework presents two key limitations. First, regarding the dependency on expert quality, the core strength of B-DPO is also its potential weakness. By design, B-DPO strongly penalizes deviations from the chosen expert trajectory $y_w$. While this enforces physical fidelity, it also risks overfitting to suboptimal or noisy expert demonstrations. In such cases, the model may fail to discover potentially more efficient or robust strategies. Second, concerning the sim-to-real gap, while SVLL's history-free stage establishes robust spatial grounding in simulation, transferring these visual policies to real-world robotic platforms involves handling unstructured noise and lighting variations. The current pixel-level alignment strategy may require additional domain randomization or perception augmentment techniques to ensure broader generalization across real-world conditions.

\section{Conclusion}
\label{sec:conclusion}
This paper introduces SVLL, a unified three-stage training framework that aligns VLM training with the physical and causal demands of embodied task planning. In the first two stages, SVLL decouples spatial grounding from temporal reasoning, thereby mitigating the risk of premature temporal binding. In the third stage, it introduces B-DPO to overcome the likelihood displacement by anchoring the policy to expert trajectories while suppressing confident hallucinations. Evaluated on the AI2-THOR simulator and real-world robotic platform, our 7B-parameter SVLL framework outperforms state-of-the-art open- and closed-source baselines in task success, yielding safer, physically-grounded agents.

% ---------------------------------------------------------------
% Bibliography
{
    \small
    \bibliographystyle{ieeenat_fullname}
    \bibliography{main} % 确保你的 .bib 文件名为 main.bib
}

\end{document}